\documentclass[letterpaper, 10 pt, conference]{ieeeconf}  

\IEEEoverridecommandlockouts                              

\usepackage{graphics} 
\usepackage{epsfig} 
\usepackage{amsmath} 
\usepackage{amssymb}  
\usepackage{hyperref}
\usepackage{cite}
\usepackage{booktabs}

\usepackage[ruled,vlined,linesnumbered]{algorithm2e}
\newcommand{\calA}{{\cal A}}

\newcommand{\calD}{{\cal D}}

\newcommand{\calU}{{\cal U}}

\newcommand{\bfc}{\mathbf{c}}

\newcommand{\bfu}{\mathbf{u}}

\newcommand{\bfx}{\mathbf{x}}
\newcommand{\bfy}{\mathbf{y}}
\newcommand{\bfz}{\mathbf{z}}

\newcommand{\bfI}{\mathbf{I}}

\newcommand{\bbE}{\mathbb{E}}

\newcommand{\bbP}{\mathbb{P}}

\newcommand{\bbR}{\mathbb{R}}

\newcommand{\ourmethod}{FogGuard}

\title{\LARGE \bf FogGuard: guarding YOLO against fog using perceptual loss}

\author{Soheil Gharatappeh, Sepideh Neshatfar, Salimeh Yasaei Sekeh$^{1}$ and Vikas Dhiman$^{2}$%
    \thanks{$^{1}$ School of Computing and Information Science, University of Maine, Orono, ME, United States,
        {\tt\small soheil.gharatappeh@maine.edu}
        }
    \thanks{$^{2}$ Department of Electrical and Computer Engineering, University of Maine, Orono, ME, United States}%
    \thanks{This material is based upon work supported by the National Science Foundation under Grant No 2218063}
}

\begin{document}
\maketitle

\begin{abstract}

In this paper, we present FogGuard, a novel fog-aware object detection network designed to address the challenges posed by foggy weather conditions. Autonomous driving systems heavily rely on accurate object detection algorithms, but adverse weather conditions can significantly impact the reliability of deep neural networks (DNNs).

Existing approaches include image enhancement techniques like IA-YOLO and domain adaptation methods. While image enhancement aims to generate clear images from foggy ones, which is more challenging than object detection in foggy images, domain adaptation does not require labeled data in the target domain. Our approach involves fine-tuning on a specific dataset to address these challenges efficiently.

FogGuard compensates for foggy conditions in the scene, ensuring robust performance by incorporating YOLOv3 as the baseline algorithm and introducing a unique Teacher-Student Perceptual loss for accurate object detection in foggy environments. Through comprehensive evaluations on standard datasets like PASCAL VOC and RTTS, our network significantly improves performance, achieving a 69.43\% mAP compared to YOLOv3's 57.78\% on the RTTS dataset. Additionally, we demonstrate that while our training method slightly increases time complexity, it doesn't add overhead during inference compared to the regular YOLO network. \footnote[3]{The code for our work is available at https://github.com/Sekeh-Lab/FogGuard.}

\end{abstract}

\section{Introduction}


Adverse weather conditions such as rain, snow, and fog present risks for driving. One such risk is reduced visibility, which, in autonomous driving, impairs object detection.
This is highly dangerous; objects that are not spotted cannot be avoided, while objects that are inaccurately localized or classified can cause the vehicle to respond by swerving or ``phantom braking"~\cite{hawkins2022VergePhantomBraking}.
In this work, we focus on improving object detection in foggy weather.

We focus on improving object detection accuracy using only cameras.
Not all autonomous vehicles have multiple sensor types, but cameras are present on virtually all of them \cite{bellan2021TechCrunchVisionOnly,bellan2022TechCrunchBackToRadar}.
This makes our research widely applicable, including to vehicles that have additional sensor types; 
camera-based object detection can always be combined with other systems to improve overall accuracy via multi-sensor fusion~\cite{bijelic-2019-seein-throug}.
Other research has explored the use of fog-specific supplemental sensors, such as the novel millimeter-wave radar~\cite{guan2020CVPRmmradar, paek2022NeuRIPSKRadar}.


The image processing community has explored the problems of dehazing, defoggification, and image-enhancement before the success of deep learning based approaches \cite{anjana2015color, lavania2012image, li2015low, hu2018exposure}. 
Bringing image processing based approaches into the learning domain, IA-YOLO~\cite{liu-2021-image-adapt} combines an image processing module with a learning pipeline to infer a de-fogged image before feeding
it into a regular object detector like YOLO~\cite{redmon-2018-yolov}. 
We posit that inferring a de-fogged image is a much harder problem than detecting objects in a foggy image. 
Clearly, detecting and classifying a bounding box in a foggy image as an object class, for example car, is a much easier problem than recreating every pixel of that car.
Additionally, dehazzing based approaches often suffer from significant computational overhead in order to achieve better image quality.

To improve object detection in a foggy image, we modify the training process of a YOLO-v3~\cite{redmon-2018-yolov} network  to be robust to foggy images.
Our modified training process \textbf{contributes} two novel ideas, 1) generalization of  perceptual loss~\cite{johnson-2016-percep-losses} to \emph{Teacher-student perceptual loss} (Section~\ref{sec:perceptual-loss}) and  2) data-augmentation with depth-aware realistic fog (Section~\ref{sec:realistic-fog}).
We use perceptual loss based on the intuition that the semantic information in a foggy image is same as that of a clear image. So we seek to minimize
the perceptual loss between a clear image and the foggified version of that iamges.
Data-augmentation is necessary because foggy object detection datasets like RTTS~\cite{li2019benchmarking} ($\sim$3K images) are much smaller than of clear image  datasets like PASCAL VOC~\cite{everingham2010pascal}($\sim$16K images) and MS-COCO~\cite{lin2014microsoft} ($\sim$116K images). 
Our ablation studies demonstrate the utility of each of our contribution in improving the accuracy of object detection in the presence of fog.

We evaluate and compare our proposed method on RTTS dataset against
state-of-the-art approaches like IA-YOLO~\cite{liu-2021-image-adapt}, 
DE-YOLO~\cite{Qin_2022_ACCV} and SSD-Entropy~\cite{bijelic-2019-seein-throug}. We find that our approach is more accurate than IA-YOLO by $11.64\%$ and \cite{bijelic-2019-seein-throug}  by $14.27\%$ while being faster by a factor of 5 times.


\section{Related Work}

%
%

Vanilla object detection algorithms~\cite{redmon-2018-yolov,bochkovskiy2020yolov4,liu-2016-ssd} are often insufficient in adverse weather conditions such as fog, rain, snow, low light scenarios.
To address such problems, the literature can be categorized into four main categories, 1) analytical image processing techniques, 2) learning-based approaches, 3) domain adaptation and 4) learning-based image-enhancement techniques.

\begin{figure*}[t]
  \includegraphics[width=\linewidth]{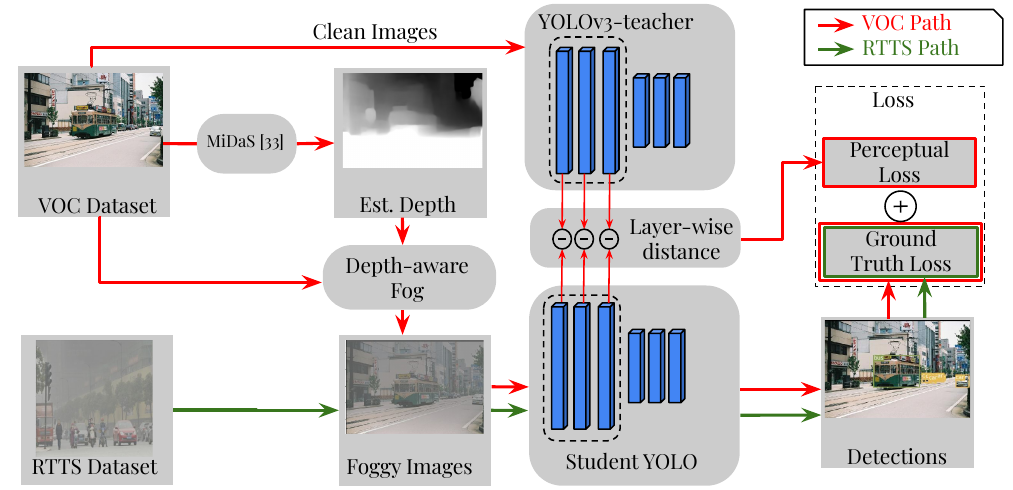}
  \caption{Training \ourmethod{} using teacher-student perceptual loss. First we train a teacher network. Then we use the teacher to train the student network using a combination of synthetic fog images and real fog images. For clear images from VOC dataset, we add synthetic depth-aware fog. Then we use the perceptual loss between the teacher and student network so that the semantic difference between the clear image and synthetic foggy images is minimized. Real foggy images are used for usual training process. This produces a student network with exception robustness to fog in real images, and high mAP on RTTS dataset.}
  \label{fig:fogrob}
\end{figure*}

\subsubsection{Analytical image processing techniques}

Analytical image processing techniques have been used to improve object detection by enhancing the quality of image.
For example, Yu et al.~\cite{1419470} introduced an adaptive approach for adjusting contrast in medical images, which can also be applied to improve object detection in low light conditions.
Zhang et al.~\cite{4476197} presented a method that effectively manipulates edge slopes to enhance sharpness and reduce noise, which can be beneficial in foggy or snowy environments.
Similarly, the dark channel prior~\cite{kaiming_2011_single_image} for image dehazing, has been used by
Chen et al.~\cite{chen2016robust} to recovery images for object detection.
However, these techniques, while effective in enhancing specific aspects of images, still fail to handle scenes with heavy fog. 
In addition, they heavily rely on physical models that require precise scene information, such as atmospheric light and scattering coefficients, making them rely on hyper-parameter tuning or estimation which has given way to learning based approaches.

\subsubsection{Learning based image-enhancement approaches}
With the success of learning algorithms, several learning-based image-enhancement approaches have been proposed.
Learning based methods have been used to find the correct image-processing filters to apply with hyper-parameter estimation, for dehazing~\cite{hu2018exposure,Yang_2018_ECCV}, improving lighting conditions~\cite{yu2018deepexposure} and improving the color and tone of the image~\cite{zhang2022deep}.
These methods have been further improved by using multi-scale convolutional neural networks (CNN)~\cite{ren2016single}, residual blocks~\cite{zhang2020drcdn} or multi-scale U-Net for dehazing~\cite{9156921}.

Two highly-cited and recent methods in this line of research are IA-YOLO~\cite{liu-2021-image-adapt} and DE-YOLO~\cite{Qin_2022_ACCV}.
IA-YOLO~\cite{liu-2021-image-adapt} uses a 5-layer CNN to estimate the hyperparameters of a differentiable image processing (DIP) module that feeds a dehazed image into an object detection algorithms, specifically YOLO-v3~\cite{redmon-2018-yolov}.
This DIP module consists of the defogging filter, white balance, gamma, tone and sharpen filtering.
Similarly, DE-YOLO~\cite{Qin_2022_ACCV}, use ``Laplacian
pyramid to decompose the input image into a low-frequency (LF) component and several high-frequency (HF) components"~\cite{Qin_2022_ACCV}. The low-frequency component (assumed to be independent of fog) is uses to enhance the high-frequency components via residual block. These different frequency components are combined into a reconstructed enhanced image. This enhanced image is fed into an object-detection algorithm like YOLOv3~\cite{redmon-2018-yolov} for improved object detection.

Image enhancement approaches, both physics-based and learning-based, seek to solve a harder problem, especially under heavy fog. For example, imagine a car in a fog where just two fog lights of the car are visible. To reconstruct an image of a car under fog, is a harder problem than to merely detect the presence of a car. To reconstruct a car, you need the color, the brand and the shape of the car. To detect a car, just the expected position of fog lights at the correct height from the road might be enough. As a result, these algorithms are often produce noisy images and their results are not reliable outputs.  

\subsubsection{Domain Adaptation}
The idea of unsupervised domain adaptation by backpropagation, developed by \cite{ganin2015UDAbyBackpropagation}, has been applied to object detection in foggy images.
The problem of domain adaptation is as follows.
Consider a source domain (for example, object detection in clear images) and a target domain (object detection in foggy images).
Assume that we have ground truth labels for the source domain, but not for the target domain.
Domain adaptation seeks to find an approach to train models for the target domain by employing similarity with the source domain.
Ganin et al.~\cite{ganin2015UDAbyBackpropagation} pose domain adaptation as a minimax problem between two competing networks, a feature extractor and a domain classifier. The domain classifier seeks to separate source and target domain inputs, while the feature extractor tries to make them indistinguishable. The features from feature extractor are also used to estimate the source domain labels.
Several works have applied domain adaptation techniques to object detection in foggy images.
Hnewa et al. \cite{9506039} study the domain shift between the distributions of the source data and proposed multiple adaptation path introduced by the domain classifiers at different scales. 
Zhou et al. \cite{ZHOU2023103649} proposed a semi-supervised domain adaptation solution for object detection, where they employed a knowledge distillation approach to obtain the features of unlabeled target domain. 
Our approach is closely related to domain adaptation, yet different. In domain adaptation, target labels are not given. For object detection in foggy images, we have labelled datasets like RTTS available. Although a domain adaptation like approaches would also need RTTS images to be labelled with the amount of fog in each image which it not available in the RTTS dataset. 
Although datasets like RTTS are much smaller than MS-COCO~\cite{lin2014microsoft} and Pascal VOC~\cite{everingham2010pascal}, we can make use of these datasets for fine-tuning.

To summarize, unlike domain adaptation methods, we do make use of target domain labels. But like domain adaptation, we enforce feature similarity between source and target domain by synthetic fog generation and teacher-student perceptual loss function.



\section{Problem formulation and Notation}
We are given two labelled datasets for object detection, a clear image dataset, $\calD_c = \{(\bfx_1^c, \bfy^c_1), (\bfx_2^c, \bfy^c_2), \dots, (\bfx_n^c, \bfy^c_n) \}$ and a foggy image dataset $\calD_f = \{(\bfx_1^f, \bfy^f_1), \dots (\bfx^f_m, \bfy^f_m)\}$. The two datasets are i.i.d. sampled from two separate and unknown distributions $\bbP_c$ and $\bbP_f$, respectively. The foggy image
dataset is much smaller than the clear image dataset $|\calD_f| \ll |\calD_c|$.
We also assume access to a (approximate) synthetic fog generator $g()$, that converts a clear image into a foggy image, $\hat{\bfx}^{cf}_i = g(\bfx^c_i;\beta)$, where $\hat{\bfx}^{cf}_i$ is synthesized foggy image from a given clear image $\bfx^{c}_i$ and a fog density parameter $\beta$.
The rest of the problem formulation is the same as object detection problem formulated by Liu et al.~\cite{liu-2016-ssd}, where the input is a color image $\bfx \in \bbR^{h \times w \times 3}$. The labels are  \(\bfy \in \bbR^{S^2B(5+C)}\) where we have $B$ bounding box labels per cell in a $S\times S$ grid formed on the image. Each bounding box label is $5+C$ dimensional, where $C$ is the number of classes.
The bounding box label vector includes the 2D position and 2D dimensions of the bounding $\bfy_{b} \in \bbR^{S^2B4}$, a one-hot vector for class probabilities $\bfy_{c} \in \{0, 1\}^{S^2BC}$, and the indicator variable $\bfy_{o} \in \{0, 1\}^{S^2B}$ for the presence or the absence of each bounding box. 
The object detection's objective is to estimate the bounding box for a given test image with a high accuracy.
The problem of object detection is to find a function that estimates all the bounding boxes for an image as correctly as possible. Given a parametric object detection model $\hat{\bfy} = F(\bfx; \Theta)$, we seek to find the parameters $\Theta$ that minimize the expected object detection loss when the input image is a foggy image,
\begin{multline*}
\small
\Theta^* = \arg\,\min_{\Theta} \mathop{\bbE}_{\small(\bfx^f,\bfy^f) \sim \bbP_f} 
\Big[ 
 \underbrace{l_{\text{loc}}(\bfy^f_b, \hat{\bfy}^f_b) 
+  
\lambda_1 l_{\text{conf}}(\bfy^f_c, \hat{\bfy}^f_c)}_{l_{\text{objdet}}}
\Big]
\end{multline*}
where the localization loss $l_{\text{loc}}$ and confidence loss $l_{\text{conf}}$ are defined as~\cite{liu-2016-ssd},
\begin{align}
    l_{\text{loc}}(\bfy^f_b, \hat{\bfy}^f_b)  &= 
    \frac{1}{b}\|(\bfy^f_b -  (P\otimes \bfI_4)\hat{\bfy}^f_b)^\top (\bfy^f_o \otimes \mathbf{1}_4)\|_\text{Huber} \notag\\
    l_{\text{conf}}(\bfy^f_c, \hat{\bfy}^f_c)  &= 
    - {(\bfy^f_c \odot (\bfy^f_o \otimes \mathbf{1}_c))}^\top \log((P \otimes \bfI_c)\hat{\bfy}^f_c)\notag\\
    &\quad- 2{\bfy^f_o}^\top \log(P\hat{\bfy}^f_o) 
    + \log(P\hat{\bfy}^f_o). \notag
\end{align}
Here $P\in \{0, 1\}^{S^2B\times S^2B}$ is a permutation matrix that greedily matches the predicted bounding boxes to the predicted bounding boxes; $b = \|\bfy^f_o\|_0$ is the number of true bounding boxes in the image; $\bfI_c$ is an identity matrix of size $c$; $\mathbf{1}_c$ is column vector of c ones; $\otimes$ is the Kronecker product; $\odot$ is element-wise product; $\lambda_1 \in [0, 1]$ is a hyper-parameter, $\|\bfz\|_\text{Huber}$ is the sum of element-wise Huber loss.




\section{Methodology}

Our method is based on two main insights.
First, the semantic distance between foggy images and clear images must be minimal, and second, all available datasets must be used to train the method.
Our method is summarized in Algorithm~\ref{alg:fogrob}.
Here we present a step-wise build-up of our method from mere fine-tuning on the foggy dataset to using fine-tuning along with perceptual loss. 

A naive approach to this problem is to use both datasets, clear $\calD_c$ and foggy $\calD_f$ and bring $\calD_c$ closer to $\calD_f$ using the synthetic fog generation function $g()$.
Let the clear image dataset be perturbed by random amount of fog $\beta_i \sim \calU[0, 0.15]$, $\calD_{g(c)} = \{ (g(\bfx^c_1; \beta_1), \bfy^c_1), \dots, (g(\bfx^c_n; \beta_n), \bfy^c_n \}$. 
The hybrid dataset is a union of the perturbed clear dataset and the foggy dataset, $\calD_{g(c)f} = \calD_{g(c)} \bigcup \calD_f$.
\begin{align}
    \Theta^* = \arg\,\min_{\Theta}& \,\, L_\text{objdet}(\calD_{g(c)f}; \Theta)
    \quad\text{ where }&\notag\\
    L_\text{objdet}(\calD_{g(c)f}; \Theta) =
    & \sum_{(\bfx_i, \bfy_i) \in \calD_{g(c)f}} \frac{l_{\text{objdet}}(\bfy_i, F(\bfx_i; \Theta))}{|\calD_{g(c)f}|}.
    \label{eq:naive-clear-fog}
\end{align}%
However, this approach depends on the synthetic fog $g()$ to be very realistic.
In other words, this approach does not force the network to explicitly ignore the effects of synthetic fog. 
To address this limitation, we take a teacher-student learning approach. 
First, we train a network on the hybrid clear-foggy dataset $\calD_{cf} = \calD_c \bigcup \calD_f$ as the teacher network, 
\begin{align}
    \Theta^*_\text{teach} = \arg\,\min_\Theta L_\text{objdet}(\calD_{cf}; \Theta).
    \label{eq:teach-loss}
\end{align}
Second, we add teacher-student loss term $l_\text{teach-stu}$ to \eqref{eq:naive-clear-fog} so that the student network estimates the same detection on a foggy image as the teacher network detects on a clear image,
\begin{multline}
    \Theta^*_{\text{stu,n}} = \arg\,\min_{\Theta} 
    L_\text{objdet}(\calD_{g(c)f}; \Theta) \\
    + \frac{\lambda_2}{|\calD_c|}\sum_{\bfx^c_i \in \calD_c} 
     \underbrace{\| F(g(\bfx^c_i); \Theta) - F(\bfx^c_i; \Theta^*_{\text{teach}}) \|_2^2}_{l_\text{teach-stu}}.
    \label{eq:activation-only-teacher-student}
\end{multline}
This loss function has the limitation that the coupling between the teacher and the student network is reduced only to the output of the last layer. 
To further couple teacher and student networks, we take inspiration from Perceptual loss~\cite{johnson-2016-percep-losses} and generalize the teacher-student coupling across multiple layers.

\subsection{Teacher-student perceptual loss}
\label{sec:perceptual-loss}
Let the layers of a $L$ layer network $F(\bfx; \Theta)$ be denoted by $f_1, f_2, \dots, f_L$, so that $F(\bfx;\Theta) = f_L(\dots f_2(f_1(\bfx)))$. Let $F_l(\bfx; \Theta) = f_l(\dots f_2(f_1(\bfx)))$ denote the network from layer 1 to layer $l$.
Perceptual loss~\cite{johnson-2016-percep-losses} between two images $\bfx_i$ and $\bfx_j$ using a pretrained network $F(\bfx; \Theta^*)$ is defined as the different between activations of given start layer $l_s$ and end layers $l_e$
\begin{align}
    l_{\text{perc}}(\bfx_i, \bfx_j) = 
    \sum_{l = l_s}^{l_e} 
    \frac{\|F_l(\bfx_i; \Theta^*) - F_l(\bfx_j; \Theta^*)\|_2^2}
    {(l_e - l_s)a_l},
\end{align}
where $a_l$ are the number of activations in the layer $l$.
Perceptual loss has been used to encode the semantic distance between two images. 
It offers a significant improvement in several tasks, like style transfer,
by focusing on high-level features of the deep networks' layers, promoting a better understanding of images~\cite{johnson-2016-percep-losses}.

We generalize the perceptual loss to a teacher-student perceptual loss and use it to penalize the semantic-distance between clear images $\bfx^c_i$ and corresponding synthetic foggy images $g(\bfx^c_i)$,
\begin{multline}
l_{\text{ts-perc}}(\bfx^c; \Theta, \Theta^*_\text{teach}) = \\
\sum_{l = l_s}^{l_e} 
\frac{\| F_l(g(\bfx^c); \Theta) - F_l(\bfx^c; \Theta^*_{\text{teach}}) \|_2^2}
{(l_e - l_s) a_l},
\label{eq:ts-perc-loss}
\end{multline}
Note that $l_\text{teach-stu}$ is special case of $l_\text{ts-perc}$ when $l_s = l_e = L$. Our joint objective function thus becomes,
\begin{multline}
    \Theta^*_\text{stu} = \arg\,\min_\Theta L_\text{objdet}(\calD_{g(c)f}; \Theta)\\
    + \frac{\lambda_2}{|\calD_c|} \sum_{\bfx^c_i \in \calD_c} l_{\text{ts-perc}}(\bfx^c_i; \Theta, \Theta^*_\text{teach}).
    \label{eq:ts-perc-objdet}
\end{multline}
Teacher-student Perceptual loss penalizes the student network when it fails to treat the foggy image
the same way as the teacher network treats a clear image.
In other words, we encourage the network to cancel the effects of fog and penalize the filters
that correspond to those features that detect fog.

\begin{algorithm}
\caption{\ourmethod{}}\label{alg:fogrob}
\SetKwInOut{Input}{In}\SetKwInOut{Output}{Out}
\SetKw{continue}{continue}
  \Input{Clear image dataset $\calD_c$, \\
  Foggy image dataset $\calD_f$\\
  The teacher network architecture $F(\bfx, \Theta)$\\
  Synthetic fog image generator $g(\bfx^c)$\\
  Optimization algorithm $\calA$ (e.g. SGD)\\
  Hyper-parameters: $T_s$ (epochs),\\
  Hyper-parameters: $\lambda_1, \lambda_2$, $l_s$, $l_e$}
  \Output{$\Theta^*_\text{stu}$}
  \BlankLine
  Train $\Theta^*_\text{teach}$ on $\calD_{cf}$\;
  Initialize $\Theta_\text{stu} \leftarrow \Theta^*_\text{teach}$\;
  \For{$t \leftarrow 0$ \KwTo $T_s$}{
    Uniformly sample $(\bfx_i, \bfy_i) \sim \calD_c \bigcup \calD_f$\;
    Compute $\hat{\bfy}_i = F(\bfx_i; \Theta_\text{stu})$\;
        \If(\tcp*[h]{Real foggy image}){$\bfx_i \in \calD_f$}{
              Update $\Theta_\text{stu}$ using $\calA$, and loss 
              $l_{\text{objdet}}(\bfy_i, \hat{\bfy}_i)$\;
              \continue\;
        }
        $\beta \sim \calU[0.0, 0.15]$ \tcp*{Intensity of fog}
        Compute depth img $d(\bfx^c)$ using MiDaS~\cite{ranftl2020MiDaS}\;
        \tcp*[h]{Generate foggy image}\;
        Compute $g(\bfx_i; \beta)$ using \eqref{eq:haze} \;
        Update $\Theta_\text{stu}$ using $\calA$, 
              and loss 
              $\lambda_2 l_\text{ts-perc}(\bfx_i, \Theta_\text{stu}, \Theta_\text{teach})$ 
              $+l_{\text{objdet}}(\bfy_i, \hat{\bfy}_i)$\;
  }
  $\Theta^*_\text{stu} = \Theta_\text{stu}$\;
\end{algorithm}

\subsection{Depth-aware realistic fog}
\label{sec:realistic-fog}
As mentioned in the problem formulation, we assume access to a synthetic fog generator $g(\bfx^c_i;\beta)$.
We choose the following synthetic fog generation model is commonly used in literature~\cite{liu-2021-image-adapt},
\begin{equation} \label{eq:haze}
    g(\bfx^c;\beta)[\bfu] = t(\bfu;\beta)\bfx^c[\bfu] + (1 - t(\bfu;\beta)) A(\bfu)
\end{equation}
where $g(\bfx^c;\beta)[\bfu]$ is the foggy image intensity at pixel $\bfu \in \bbR^2$, $\bfx^c[\bfu]$ is the clear image intensity at pixel $\bfu$,
$A(\bfu)$ is the atmospheric light (fog), and $t(.)$ is the transmission, the portion of the
light that is not scattered due to fog and has reached to the camera. 
Note that $A(\bfu)$ can be used to generate different adverse weather patterns, such as snow and rain, apart from fog. In this work, we focus on fog, and leave snow and rain for future work. For fog generation $A(\bfu)$, we use $A=0.5$ as a constant.
The transmittance $t(.)$ decreases exponentially with depth $d(\bfu)$,
\begin{equation}
  \label{eq:transmission}
  t(\bfu;\beta) = \exp(-\beta d(\bfu)),
\end{equation}
when the fog is homogeneous and $\beta$ is a measure of fog density.
The first term $t(\bfu;\beta)\bfx^c[\bfu]$ is called direct attenuation and describes the scene radiance and its decay in the medium. 
The second term $(1 - t(\bfu;\beta)) A(\bfu)$ is called airlight, which represents
the shift to the scene colors that is caused by previously scattered light.

Computing foggy image $g(\bfx^c;\beta)$, depends on the depth information $d(\bfu)$, but none of the datasets RTTS~\cite{li2019benchmarking}, MS-COCO~\cite{lin2014microsoft} or PASCAL-VOC~\cite{everingham2010pascal} used for object-detection training provide depth.
Methods like IA-YOLO~\cite{liu-2021-image-adapt}, instead use a pseudo-depth model with maximum depth at the center and depth radially decreasing towards the edges,
\begin{align}
d_{IA}(\bfu) =  \sqrt{\max(W, H)} - 0.04 \|\bfu - \bfc\|,
\label{eq:pseudo-depth}
\end{align}
where $W$ is the width of the image, $H$ is the height and $\bfc = [W/2, H/2]$ is the center of the image.
In this work, to estimate depth, we use a neural network model, MiDaS~\cite{ranftl2020MiDaS}, that estimates depth from a single-image. 
We show the fog rendered by this method in Figure~\ref{fig:fogrob}, which visually appears realistic. 
We also show the importance of using this realistic depth in ablation studies (Section~\ref{sec:ablation-realistic-depth}).
%

\section{Experiments}

We evaluate our proposed method, \ourmethod{}, with extensive set of experiments.
Before describing the results of our experiments, we discuss the common aspects of the experimental setup, including the chosen object detection datasets, implementation details, and the evaluation metric.

\subsection{Datasets}

We mainly depend on two datasets, the clear image dataset $\calD_c$ as PASCAL VOC (16,552 images) and the foggy image dataset $\calD_f$ as RTTS (3,026 images).
We create a hybrid dataset $\calD_{cf}$ by combining the two.
Then the hybrid dataset is divided into 80-20 training-validation split achieved through sampling with replacement.

Finally, to evaluate we set aside, 4952 images from PASCAL VOC and 432 from RTTS, as a test dataset.

For some of the experiments,  we also use three variations of the PASCAL VOC dataset where 
replace all the images with synthetic fog using $g(\bfx^c_i; \beta)$. 
We call the three variations of dataset $\calD_{g(c,\beta)f}$ as \emph{LowFog}, \emph{MediumFog} and \emph{HeavyFog} for $\beta = \{0.05, 0.10, 0.15\}$, respectively.



\subsection{Implementation Details}
We use a YOLO-v3 network as the main backbone neural network architecture.
We initialize the network weights pre-trained on MS-COCO, which is an even larger dataset for object detection.
Training the \ourmethod{} starts with training the Teacher.
The teacher network $\Theta^*_\text{teach}$ is trained by Stochastic Gradient Descent (SGD) optimizer for $600$ epochs on the hybrid dataset $\calD_{g(c)f}$.
The learning rate is set to $0.001$ and the batch size is $16$. The Student network, aka. \ourmethod{}, has identical specifications, and uses the same dataset to be trained.
We use Pytorch to implement our network and run the experiments on an A100 GPU. All experiments are repeated 3 times with reported mean and standard deviation as the errors.

%

\subsection{Evaluation metric}
The main metric we use is the mean Average Precision (mAP) as described in Russakovsky et al.~\cite{russakovsky2015imagenet}.
To compute mAP, we follow a two step process. First, for a given threshold $t \in [0, 1]$, all the detected bounding boxes are greedily matched to the true bounding boxes~\cite[Alg.2]{russakovsky2015imagenet}. 
The two bounding boxes are considered a match of the intersection over a union of the two bounding boxes is greater than the threshold $t$.
The mAP is computed as the ``average \emph{precision} over the different levels of \emph{recall} achieved by varying the threshold $t$''~\cite{russakovsky2015imagenet}; which is further averaged over all the object classes.
Here, \emph{recall} is the ratio of the number of correct bounding boxes to the number of true bounding boxes.
\emph{Precision} is the ratio of the number of correct bounding boxes to the number of predicted bounding boxes.

\section{Results}

We discuss the results of comparing our methods on three categories of experiments, 1) comparison against the state-of-the-art baselines, 2) ablation studies, and 3) efficiency analysis.

\subsection{Comparison against baseline methods}

We compare our method against four state-of-the-art baseline methods: IA-YOLO~\cite{liu-2021-image-adapt}, DE-YOLO~\cite{bijelic-2019-seein-throug}, and  SSD+Entropy on both the datasets, (1) RTTS~\cite{li2019benchmarking} and (2) VOC~\cite{everingham2010pascal}. 

\paragraph{IA-YOLO}
IA-YOLO~\cite{liu-2021-image-adapt} is evaluated on both PASCAL VOC and RTTS
datasets, in which the trained network was provided by the author here. In order to
make our comparison more fair, we fine-tuned the network on RTTS data for $50$
more epochs and reported the best performance across both datasets.

\paragraph{DE-YOLO}
In our experiments, we tested the DE-YOLO method using the provided checkpoint for the network. We evaluated its performance on the test set of VOC dataset and RTTS dataset without performing any additional training, as the training script was not provided by the authors.

\paragraph{SSD+Entropy}
This approach is developed upon the foundation established in Bijelic et al.~\cite{Bijelic_2020_STF} which incorporates Entropy uncertainty within the SSD~\cite{liu2016ssd} architecture. However, in contrast to their study that utilized multi-modal inputs, our baseline focuses on leveraging single-modal image inputs of VOC and RTTS datasets.


In Table~\ref{tab:od-baseline}, we summarize the mAP(\%) of our method compared against baselines on RTTS and PASCAL VOC. 
Our method beats all the baselines considered, IA-YOLO, DE-YOLO, and SSD-Etropy on both RTTS and VOC.
\begin{table}[h!]
  \centering
  \begin{tabular}{ l   r   r   r   r   r }
   \toprule
    Eval. D.   &  \ourmethod{}     & IA-YOLO~\cite{liu-2021-image-adapt} & DE-YOLO~\cite{Qin_2022_ACCV}     &    SSD$+$Ent  \\
   \midrule
  \textbf{RTTS} 		  &  \textbf{69.43}  & 57.78   & 55.16    &            42.92          \\
  VOC 		  &  \textbf{92.22}  & 71.62   & 81.60    &    69.4          \\
   \bottomrule
   \end{tabular}
\caption{Comparing \ourmethod{} with prominent object detection algorithms using Mean Average Precision.}
\label{tab:od-baseline}
\end{table}

\subsection{Ablation Studies}

We do five ablation studies to understand the importance of different components of our pipeline.
%
%

\subsubsection{Comparison against the teacher network}
We compare our method against a teacher YOLOv3 network trained on the hybrid dataset $\calD_{cf}$ using~\eqref{eq:teach-loss}.
We present the mAP(\%) of our each network evaluated on VOC, RTTS, Mixture perturbed with \emph{LogFog}, \emph{MediumFog} and \emph{HeavyFog}.

The first (\ourmethod{}) and second column (\emph{YOLOv3-teacher}) of Table~\ref{tab:ablation} illustrates the superior performance of \ourmethod{} compared to \emph{YOLOv3-teacher} evaluated across different datasets. 
This highlights the importance of using sythentic fog and incorporating
perceptual loss in the training process. 
Especially noteworthy is the clear superiority of \ourmethod{} over
the YOLOv3-teacher model, as evidenced by the substantial performance gap of
nearly $5\%$ observed in the case of the RTTS dataset.

\begin{table*}
  \centering
  \begin{tabular}{ l c c c c c }
   \toprule
    Eval Dataset          & \ourmethod{}          & YOLOv3-teacher        & w.o. Perc. Loss        & w. RTTS                  & IA-YOLO Depth          \\
   \midrule                
  RTTS                    & \textbf{69.84} ($\pm$ 1.51)	& 65.16  ($\pm$ 2.79)  & 62.36 ($\pm$ 3.30)     & 68.56  	($\pm$ 4.12) 	& 63.31  ($\pm$ 4.97)  	 \\
  Mixture $+$ HeavyFog   & 70.37 ($\pm$ 0.45)	& 68.68  ($\pm$ 0.35)  & 68.99 ($\pm$ 0.51)     & 70.16  	($\pm$ 0.46) 	& \textbf{70.92}  ($\pm$ 0.58)  	 \\
  Mixture $+$ MediumFog  & \textbf{70.57} ($\pm$ 0.41)	& 69.52  ($\pm$ 0.28)  & 69.95 ($\pm$ 0.43)     & 70.13  	($\pm$ 0.49) 	& 71.04  ($\pm$ 0.37)  	 \\
  Mixture $+$ LowFog     & 70.66 ($\pm$ 0.34)	& 70.26  ($\pm$ 0.38)  & 70.30 ($\pm$ 0.48)     & 70.31  	($\pm$ 0.71) 	& \textbf{70.73}  ($\pm$ 0.50)  	 \\
  VOC                     & 90.10 ($\pm$ 5.39)	& 89.11  ($\pm$ 4.35)  & 85.39 ($\pm$ 3.95)     & 91.99    ($\pm$ 1.98)	& \textbf{92.72}  ($\pm$ 0.63)    \\
   \bottomrule
\end{tabular}
\caption{Ablation studies: Variations of \ourmethod{} with and without several components for ablation studies.}
\label{tab:teacher-vs-student}
\label{tab:ablation}
\end{table*}

\subsubsection{Importance of the Teacher-Student Perceptual Loss}




In the third column of Table~\ref{tab:ablation} we demonstrate the effect of not including the
perceptual loss in the \ourmethod{} network. By removing the perceptual loss from
the network, we are left with the teacher network (\emph{w.o. Perc. Loss}) trained using the hybrid perturbed dataset, $\calD_{g(c)f}$ using loss \eqref{eq:naive-clear-fog}.  
Training \ourmethod{} without incorporating perceptual loss is
equivalent to training a vanilla YOLOv3 network with fog-based data augmentation.
Not including the perceptual loss is detrimental to the performance of our network, as the \ourmethod{} outperforms \emph{w.o Perc. Loss}) network.

\subsubsection{Including RTTS dataset in Perceptual loss hurts performance}
As mentioned in \eqref{eq:ts-perc-objdet}, the teacher-student perceptual loss is only computed
over the clear image dataset $\calD_c$.
We ask the question, whether including the foggy image 
dataset RTTS in the perceptual loss term, loses or gains performance.
The fourth column of Table~\ref{tab:ablation} shows that by including RTTS into
perceptual loss computation, we lose $1.28\%$ in RTTS evaluation and gain
$1.89\%$ in VOC. 
This shows that the \ourmethod{} is slightly more accurate when dealing with real fog, and slightly less accurate when dealing with clear images.
However, our network is generally more robust towards real fog as the
standard deviation of the network in Table \ref{tab:ablation} for RTTS dataset
is almost $3$ times higher than the network in Table \ref{tab:teacher-vs-student}.

\subsubsection{Importance of realistic depth in fog generation}
\label{sec:ablation-realistic-depth}
As discussed in Section~\ref{sec:realistic-fog}, in order to create an image with a more realistic fog effect, we need to have the image's depth information.
%
The depth information is only necessary during the training
phase of \ourmethod{}, not during the inference. 
We get the depth from MiDaS~\cite{ranftl2020MiDaS}, while IA-YOLO uses a 
psuedo-depth as desribed in \eqref{eq:pseudo-depth}.
IA-YOLO assumes the maximum depth in the center, which decreases radially outwards.
The last column in Table \ref{tab:teacher-vs-student} shows that when the synthetic fog is generated using the IA-YOLO approach, the network experiences a decline of $6.53\%$ in mAP over RTTS but increase by 2.62\% over VOC.
This finding highlights the
significance of realistic depth in enhancing the network's ability to accurately detect objects in foggy conditions. 
The increase in mAP on clear and synthetic fog images when using psuedo-depth is surprising. 
One hypothesis is that MiDaS-based fog is more complex to understand and remove than psuedo-depth fog.
This hurts the performance in clear datasets and synthetic fog images.

\subsubsection{Location and number of layers used in Perceptual loss}
Lastly, we explore the impact of the location and the number of layers considered for computing
the perceptual loss in \ourmethod{}.
The perceptual loss defined in \eqref{eq:ts-perc-loss} depends upon 
the start layer $l_s$ and end layer $l_e$. 
We evaluate two locations first few layers $l_s=1$ and the last few layers $l_e=L$. 
We also vary the number of layers $l_e - l_s$ between 4 to 9. 
The results are shown in Table~\ref{tab:layers_perc}. 
For this experiment, we fixed the number of training epochs to 300 (vs. 600 in the original experiment), and find that including a higher number of layers for perceptual
loss computation generally leads to better object detection
performance.
However, there is trade-off between the computational cost during training involved to calculate the difference between the full set of layers in the network, or only considering a subset of layers. We found that the duration of our experiment nearly triples when utilizing 9 layers instead of just 4 layers.


\begin{table}
  \centering
  \begin{tabular}{ l   c   c   c   c   c   c  }
    \toprule
     Location & \multicolumn{3}{c }{First ($l_s=1$)} & \multicolumn{3}{c}{Last ($l_e = L$)}  \\
     \cmidrule(lr){2-4} \cmidrule(lr){5-7}
    Layers ($l_e - l_s$)& 4     & 7     & 9     & 9     & 7          & 4       \\
    \midrule
    RTTS            & 66.13 & 65.12 & \textbf{73.58} & 70.85  & 66.62 & 66.62   \\ 
    HeavyFog       & 70.38 & 70.22 & 69.85 & 70.25  & 70.44 & 70.50   \\    
    MediumFog      & 70.81 & 70.51 & 70.37 & 70.43  & 70.84 & 70.75   \\ 
    LowFog         & 70.85 & 70.59 & 70.28 & 70.28  & 70.94 & 70.97   \\ 
    VOC             & 91.98 & 90.24 & 94.60 & 94.37  & 92.93 & 92.21   \\ 
    \bottomrule
  \end{tabular}
  \caption{Layers involved in computation of perceptual loss}
  \label{tab:layers_perc}
\end{table}

\subsection{Efficiency Analysis}

\ourmethod{} has zero overhead with respect to the vanilla YOLO during inference
time. The network has exactly the same quantity of parameters, with no added cost
corresponding to extra blocks same as seen in IA-YOLO
\cite{liu-2021-image-adapt}. During the training, however, our method still needs to maintain two distinct YOLO networks, namely the Teacher and Student in order to compute the perceptual loss. So, \ourmethod{} is as fast as its backbone network; YOLO-v3. The zero-overhead characteristic also allows our proposed method to be easily adapted to other scenarios where enhanced accuracy or speed is desired. 
In such cases, we can replace just the backbone YOLO-v3 network, and train an alternative network that is more robust against the adverse conditions.

In order to demonstrate the efficiency of \ourmethod{}, we evaluated both \ourmethod{} and IA-YOLO on 100 samples and measured the execution time. In this experiment, \ourmethod{} performed 5.1 times faster than IA-YOLO (139.68s vs 27.42s).
\section{Conclusion}
We proposed \ourmethod{}, a novel approach for effectively detecting objects in
challenging weather conditions.
Our approach utilizes a Teacher-Student architecture, with the student network
incorporating a perceptual loss to enhance robustness against adverse weather
conditions. By synthesizing realistic fog in the input images and emulating the
performance of the teacher network, our approach achieves remarkable
results. Through rigorous experimentation, we have demonstrated the superiority
of our optimized training method and thoughtfully designed dataset,
outperforming existing methods by a significant margin. We also showed that our
proposed method does not induce any overhead to the vanilla YOLO network, making
it a suitable approach for real-world problems such as autonomous driving.

\addtolength{\textheight}{0cm}   




\section*{ACKNOWLEDGMENT}



This work has been partially supported (Vikas Dhiman) by NSF RII Track-2 FEC 2218063 and (Soheil Gharatappeh, Sepideh Neshatfar, and Salimeh Yasaei Sekeh) by NSF CAREER-CCF 5409260; the findings are those of the authors only and do not represent any position of these funding bodies.

\bibliographystyle{IEEEtran} 

\bibliography{main}


\end{document}